\documentclass[runningheads]{llncs}
\usepackage{amsmath}
\usepackage{multirow}
\usepackage{tabularx}
\usepackage[colorlinks,citecolor=blue]{hyperref}
\usepackage{pgfplots}
\pgfplotsset{compat=1.14}

\usepackage[T1]{fontenc}
\usepackage{graphicx}
\begin{document}
\title{FLDNet: A Foreground-Aware Network for Polyp Segmentation Leveraging Long-Distance Dependencies}
\titlerunning{FGA-Net with LDD for Polyp Seg}
\author{Xuefeng Wei\inst{1}${\dagger}$, Xuan Zhou\inst{1}${\dagger}$}

\authorrunning{X. Wei et al.}

\institute{Institut Polytechnique de Paris, Rte de Saclay, 91120 Palaiseau, France 
\email{xuefeng.wei@ip-paris.fr}\\
\email{xuan.zhou@ip-paris.fr}}

\maketitle              
\renewcommand{\thefootnote}{}
\footnotetext[1]{$^{\dagger}$ Indicates equal contribution and Corresponding author.}

\begin{abstract}
Given the close association between colorectal cancer and polyps, the diagnosis and identification of colorectal polyps play a critical role in the detection and surgical intervention of colorectal cancer. In this context, the automatic detection and segmentation of polyps from various colonoscopy images has emerged as a significant problem that has attracted broad attention. Current polyp segmentation techniques face several challenges: firstly, polyps vary in size, texture, color, and pattern; secondly, the boundaries between polyps and mucosa are usually blurred, existing studies have focused on learning the local features of polyps while ignoring the long-range dependencies of the features, and also ignoring the local context and global contextual information of the combined features. To address these challenges, we propose \textbf{FLDNet} (\textbf{F}oreground-\textbf{L}ong-\textbf{D}istance \textbf{N}etwork), a Transformer-based neural network that captures long-distance dependencies for accurate polyp segmentation. Specifically, the proposed model consists of three main modules: a pyramid-based Transformer encoder, a local context module, and a foreground-Aware module. Multilevel features with long-distance dependency information are first captured by the pyramid-based transformer encoder. On the high-level features, the local context module obtains the local characteristics related to the polyps by constructing different local context information. The coarse map obtained by decoding the reconstructed highest-level features guides the feature fusion process in the foreground-Aware module of the high-level features to achieve foreground enhancement of the polyps. Our proposed method, FLDNet, was evaluated using seven metrics on common datasets and demonstrated superiority over state-of-the-art methods on widely-used evaluation measures.


\end{abstract}
\section{Introduction}
Colorectal cancer (CRC) is the third most common type of cancer worldwide, with polyps on the intestinal mucosa considered as precursors of CRC that can easily become malignant. Therefore, the early detection of polyps has significant clinical implications for the treatment of CRC. Numerous studies have shown that early colonoscopy reduces the incidence of CRC by 30\%~\cite{14}. Fortunately, due to advancements in computing, several automatic polyp segmentation techniques~\cite{2,3} have been proposed, demonstrating promising performance~\cite{4,8}. These techniques, however, still face challenges in accurately segmenting polyps due to their varied sizes and shapes, and the indistinct boundaries between polyps and the mucosa~\cite{17}.

To tackle these challenges, the application of conventional Convolutional Neural Networks (CNNs) has been explored, which possess a local receptive field. However, CNNs may neglect the global information of polyps, which could lead to reduced performance in polyp segmentation~\cite{25,31}. The Transformer model~\cite{19}, with its self-attention mechanism, is able to focus on the global context of input data. This property is particularly important when dealing with the complexity and diversity of polyp images.

Moreover, some studies have proposed methods that either leverage attention mechanisms~\cite{15} or utilize individual edge supervision~\cite{20} to handle the varying polyp sizes and shapes. But these methods tend to focus more on local information, lacking the consideration of the global context. Therefore, in this study, we adopt a Transformer-based model, which we expect can overcome the limitations of CNNs in handling polyp images and achieve higher segmentation performance.

To further address these challenges and enhance the accuracy and generalization of polyp segmentation techniques, we propose the PVT-based FANet polyp segmentation network, designed to capture long-distance dependencies between image patches. To compensate for the shortcomings of PVT~\cite{7}, we introduce a local context module and a foreground perception module. The local context module builds different context information through a multi-branch structure to obtain polyp local information, thus resolving the issue of attention dispersion and indistinct features of local small objects. The foreground perception module addresses the problem of low contrast and blurred boundaries between polyps and surrounding mucosa. By highlighting foreground features in a chaotic background, we aim to achieve more accurate segmentation.

\vspace{0.5em}  
\noindent The main contributions of this paper are: 
\begin{enumerate}
  \setlength\itemsep{1em}  
  \item We introduce a pyramidal Transformer, enhancing the model's ability to capture long-distance dependencies and boosting the model's capacity to seize global information and generalization capability.
  \item We introduce a local context module and a foreground-Aware module, addressing the deficiency of local features in polyps and the blurring of the boundaries between polyps and surrounding mucosa, thus enhancing the network's capability to capture local information. 
  \item Extensive experiments demonstrate that our proposed FLDNet surpasses most state-of-the-art models on challenging datasets.
\end{enumerate}

\section{Related Work}\label{related:work}
The field of polyp segmentation has witnessed significant developments, transitioning from relying on handcrafted features to adopting deep learning methods. Early studies heavily relied on handcrafted features such as color and texture\cite{2,3}. However, these features often struggle to capture global information and exhibit poor stability in complex scenes.

In recent years, deep learning methods have gained dominance in the field of polyp segmentation. Approaches such as Fully Convolutional Networks (FCN)\cite{4}, U-Net\cite{24}, U-Net++\cite{25}, ACS-Net\cite{30}, and PraNet\cite{15} have been applied to polyp segmentation. The U-Net model\cite{24} utilizes a well-known structure for medical image segmentation, consisting of a contracting path that captures context and an expanding path that restores precise details. U-Net++\cite{25} and ResUNet++\cite{17} further improve the original U-Net by incorporating dense connections and better pre-trained backbone networks, achieving satisfactory segmentation performance. Although these methods yield good results for the main body of polyps, the boundary regions are often neglected. To enhance boundary segmentation, Psi-Net\cite{41} proposes a method that combines both body and boundary features. Similarly, SFA\cite{29} explicitly applies region-boundary constraints to supervise the learning of polyhedron regions and boundaries, while PraNet\cite{15} introduces a reverse attention mechanism that first localizes the polyp region and then implicitly refines the object boundary.

Recently, with the advent of Transformer architectures in computer vision tasks\cite{19}, researchers have started to explore their potential in medical image analysis, including polyp segmentation. Unlike traditional convolutional layers, the Transformer model captures long-range dependencies in the data, which could be particularly beneficial in handling polyp images' complexity and diversity. However, directly applying Transformer models in medical image segmentation tasks, such as polyp segmentation, is not straightforward\cite{42}. These models were originally designed for natural language processing tasks, and their suitability for the specific challenges of medical image segmentation is still an open question. For instance, the self-attention mechanism of Transformer models, while being powerful, is also computationally expensive and could be less effective in localizing small-sized polyps due to its global nature.

Moreover, it's worth mentioning that while deep learning techniques have significantly improved polyp segmentation, issues still persist. One key challenge is the foreground-background imbalance present in most polyp images, which could affect the segmentation performance. Another challenge is the large variance in polyps' appearance, including their size, shape, texture, and color. These variances make it hard for models to generalize across different patients and imaging conditions.

To overcome these challenges, there is an increasing trend towards developing more sophisticated models that combine the strengths of different architectures. For example, hybrid models combining the local feature extraction capability of CNNs and the long-range dependencies modeling of Transformer models have been proposed.

\begin{figure}[t!]
\includegraphics[width=\textwidth]{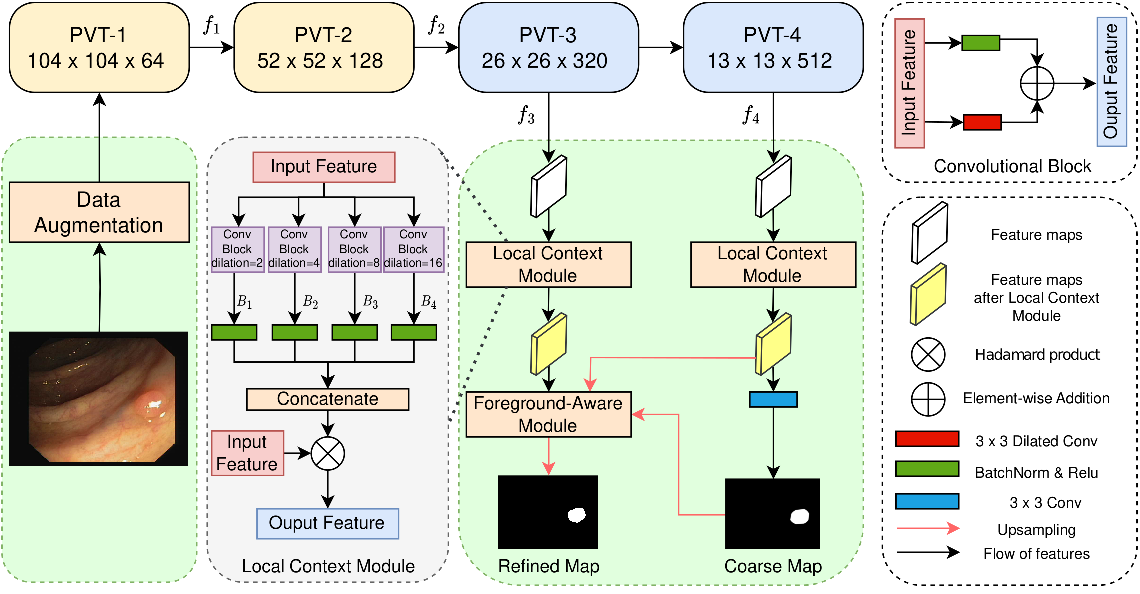}
\caption{An overview of the proposed FLDNet. The model consists of a Pyramid Vision Transformer (PVT)-based encoder, a Local Context Module, and a Foreground-Aware Module. The PVT encoder captures long-distance dependencies and forms the backbone of the model. It is divided into four sections, each corresponding to different resolution feature maps. Features from the last two blocks {${f_i}_{i=3}^{4}$} are further utilized considering their higher semantic information. The Local Context Module and the Foreground-Aware Module further refine the features for precise polyp segmentation. The specific operations and interactions between these modules are described in detail in \S\ref{sec:second}.} \label{fig1}
\end{figure}

\section{Method}\label{sec:second}
In the proposed FLDNet, we employ a PVT-based encoder, Local Context Module, and Foreground-Aware Module to achieve precise polyp segmentation (Figure \ref{fig1}). The training procedure and loss function used for network optimization are also discussed in this section.

The PVT-based encoder forms the backbone of the model, capturing long-distance dependencies. Deeper features from the PVT encoder, bearing more semantic information, are specifically employed in our approach. Further details on each component of FLDNet are provided in the ensuing discussion.

\subsection{Local Feature Capture Via Local Context Module}\label{sec:second_sub1}
CNN-based models \cite{24,25,17} commonly used in medical image segmentation can struggle to extract critical local details such as textures and contours. Our Local Context Module mitigates this by using standard and dilated convolutions to build diverse context information.

The Local Context Module, visualized in Figure \ref{fig1}, consists of four branches. Each branch involves both standard and dilated convolutions, processes the input feature map $X$, and combines the convolution outputs using element-wise addition. The outputs from all branches are then concatenated and element-wise multiplied with the original input feature map $X$ to yield the final module output $Y$. This process can be encapsulated in a single equation:

\begin{equation}
Y = X \otimes Concat(Conv_i(X) \oplus ConvDilated_{2^i}(X))_{\forall i}
\end{equation}
Where $\oplus$ represents the element-wise addition operation, $\otimes$ denotes the element-wise multiplication, and the subscript $\forall i$ signifies that the operation applies for all branches $i$. This approach allows effective capture of both global and local context information, which is pivotal for accurate polyp segmentation.

\begin{figure}[t!]
\includegraphics[width=\textwidth]{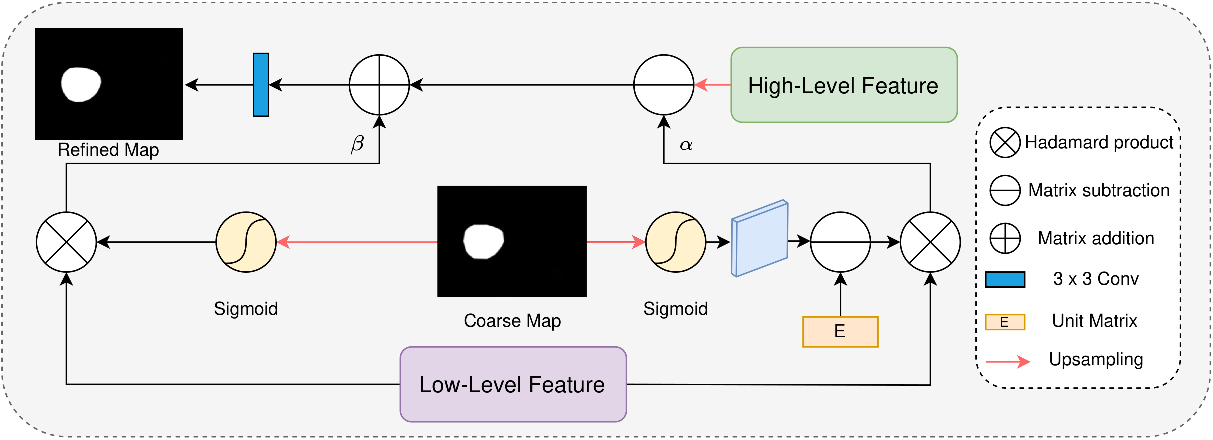}
\caption{An overview of the proposed Foreground-Aware Module, See \S\ref{sec:second_sub2} for details.} \label{fig3}
\end{figure}
\subsection{Foreground-Aware Module}\label{sec:second_sub2}
In contrast to strategies in \cite{14,6,15}, which use features from all stages for segmentation, we adaptively fuse high-level features in two parallel streams, thereby improving foreground depiction. Our method, inspired by \cite{15}, leverages features from PVT, known for their rich contextual information, instead of conventional deepest-layer CNN features.

Our Foreground-Aware Module, illustrated in Figure \ref{fig3}, employs a two-step approach: First, high-level features are decoded to generate a coarse segmentation map, guiding the fusion of deep-layer features to accentuate the foreground. Second, we derive a foreground probability map via upsampling the segmentation map and applying the Sigmoid function. A corresponding background map is computed by subtracting the foreground map from 1, enabling the separation of foreground and background features.

Next, two learnable parameters, $\alpha$ and $\beta$, are used to balance the impact of foreground and background features on the final refined segmentation map. The adjusted feature map, generated by reducing $\alpha$ times the foreground map and increasing $\beta$ times the background map, is then convolved to yield the final segmentation map. This is described by the equation:

\begin{equation}
R = {Conv}({Upsample}(H) - \alpha \cdot F' + \beta \cdot B')
\end{equation}
Where $H$ denotes the high-level features, $Conv$ the convolution operation, and $F'$ and $B'$ the foreground and background feature maps, respectively. $\alpha$ and $\beta$ are learnable parameters.

\subsection{Loss Function}\label{sec:second_sub3}
Our training objective employs a combined loss function as suggested in \cite{26}, incorporating the weighted Intersection over Union (wIoU) loss and the weighted Binary Cross Entropy (wBCE) loss. The total loss, \(L_{\text{total}}\), is expressed as:

\begin{equation}
L_{\text{total}} = L_{\text{wIoU}} + L_{\text{wBCE}} + \sum_{i=3}^{4} L(G, S_{i}^{up})
\end{equation}
In these equations, \(L_{\text{wIoU}}\) and \(L_{\text{wBCE}}\) represent the wIoU and wBCE losses respectively. \(G\) indicates the ground truth, \(S_{i}^{up}\) refers to the upscaled prediction at scale \(i\), and \(L\) symbolizes the loss function, identical to \(L_{\text{total}}\) without the summation term. The variables \(w_{i}\), \(p_{i}\), and \(y_{i}\) correspond to the weights, predictions, and ground truth at the \(i^{th}\) pixel respectively.

This formulation facilitates our model in capturing the complex task of polyp segmentation, ensuring the retention of detailed information across various scales.

\section{Experiment}
\subsection{Datasets and Baselines}\label{sec:third_sub1}
We assess the FLDNet using three polyp segmentation datasets: CVC-ClinicDB \cite{18}, CVC-ColonDB \cite{27}, and ETIS \cite{28}. CVC-ClinicDB and CVC-ColonDB offer a broad spectrum of polyps in different imaging conditions, while ETIS is distinctive for its low contrast, complex images.

\textbf{UNet} \cite{24} and \textbf{UNet++} \cite{25} are known for their effective feature capture. \textbf{SFA} \cite{29} has a shared encoder and two constrained decoders for accurate predictions. \textbf{ACSNet} \cite{30} addresses polyp size, shape, and spatial context, while \textbf{PraNet} \cite{15} uses a reverse attention mechanism for incremental object refinement. \textbf{DCRNet} \cite{31} leverages contextual relations to enhance accuracy, \textbf{MSEG} \cite{37} incorporates the HarDNet68 backbone. \textbf{EU-Net} \cite{33} combines the SFEM and AGCM to prioritize important features. Lastly, \textbf{SANet} \cite{32} uses a shallow attention module and a probability correction strategy to ensure accuracy, particularly for small polyps.
\begin{figure}[t!]
\includegraphics[width=\textwidth]{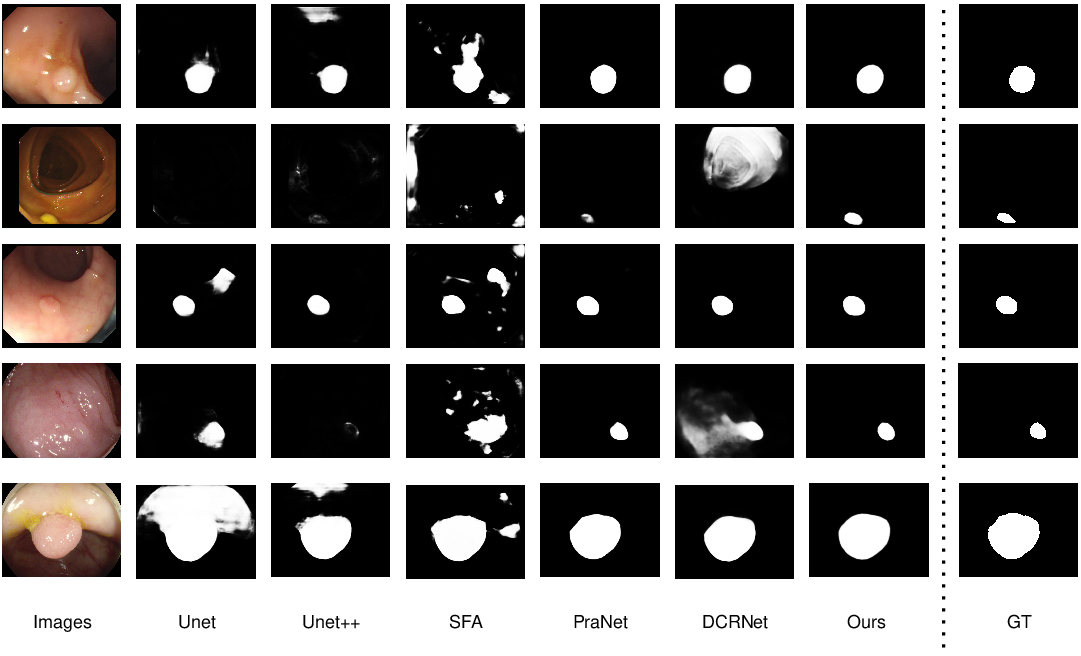}
\caption{Visual comparison between proposed FLDNet and other methods. Example polyp images sourced from the CVC-ClinicDB and ETIS datasets. The first row presents the original polyp images, followed by the corresponding segmentation results, where the polyp regions are denoted in white and the background in black. To facilitate comparison, the Ground Truth (GT) and the segmentation predictions generated by our model are separated by dashed lines.} \label{fig4}
\end{figure}

\subsection{Evaluation Metrics}\label{evaluation}

To evaluate the performance of our model, we employ several widely used metrics, including the Dice Coefficient (Dice), Intersection over Union (IoU), weighted F-measure ($F_{\beta}^{w}$), S-measure ($S_{\alpha}$), mean absolute error (MAE), and E-measure. For Dice and IoU, we report the mean values, denoted as mDice and mIoU, respectively. For the E-measure, we provide both the mean and max values, represented as $mE_{\xi}$ and $maxE_{\xi}$. These metrics are extensively utilized in the literature for segmentation and object detection tasks \cite{15}.

\subsection{Experiment Settings}\label{sec:third_es}
We normalized the input images to a size of 416$\times$416 and employed data augmentation techniques such as random flipping and brightness variations. The framework used was PyTorch 1.8.1 with CUDA 11.1, running on a hardware configuration consisting of a GeForce RTX 3090 with 24GB of compute memory. The initial learning rate was set to $10^{-4}$, and the Adam optimizer was utilized. The learning rate was reduced by a factor of 10 every 50 epochs, and a total of 200 epochs were trained.

\subsection{Quantitative Comparison}\label{sec:third_sub2}
In order to thoroughly evaluate the effectiveness of our proposed FLDNet for polyp segmentation, we conducted a comparative analysis with several state-of-the-art models. The results of this comparison on the CVC-ClinicDB dataset \cite{18}, CVC-ColonDB dataset \cite{27}, and ETIS dataset \cite{28} are detailed in Table \ref{tab:table1}.

\begin{table}[htp]
\centering
\caption{Comparison of different models on different datasets. Bold values represent the best results, and underlined values indicate the second best results .}
\label{tab:table1}
\begin{tabularx}{\textwidth}{cXXXXXXXXX}
\hline
Dataset & Baseline & mDic & mIoU & $F_{\beta}^{\omega}$ & $S_{\alpha}$ & $mE_{\xi}$ & $maxE_{\xi}$ & MAE $\downarrow$\\
\hline
\multirow{8}{*}{CVC-ClinicDB} & UNet & 0.823 & 0.755 & 0.811 & 0.889 & 0.913 & 0.954 & 0.019 \\
& UNet++ & 0.794 & 0.729 & 0.785 & 0.873 & 0.891 & 0.931 & 0.022 \\
& SFA & 0.700 & 0.607 & 0.647 & 0.793 & 0.840 & 0.885 & 0.042 \\
& ACSNet & 0.882 & 0.826 & 0.873 & 0.927 & 0.947 & 0.959 & 0.011 \\
& PraNet & 0.899 & \textbf{0.849} & \textbf{0.896} & 0.936 & 0.959 & 0.965 & \underline{0.011} \\
& DCRNet & 0.896 & 0.844 & 0.890 & 0.933 & 0.964 & \textbf{0.978} & 0.010 \\
& EU-Net & \underline{0.902} & 0.846 & 0.891 & \underline{0.936} & \underline{0.959} & 0.965 & 0.011 \\
& \textbf{Ours} & \textbf{0.905} & \underline{0.848} & \underline{0.894} & \textbf{0.937} & \textbf{0.964} & \underline{0.974} & \textbf{0.010} \\
\hline
\multirow{10}{*}{CVC-ColonDB} & UNet & 0.512 & 0.444 & 0.498 & 0.712 & 0.696 & 0.776 & 0.061 \\
& UNet++ & 0.483 & 0.410 & 0.467 & 0.691 & 0.680 & 0.760 & 0.064 \\
& SFA & 0.469 & 0.347 & 0.379 & 0.634 & 0.675 & 0.764 & 0.094
\\
& ACSNet & 0.716 & 0.649 & 0.697 & 0.829 & 0.839 & 0.851 & 0.039 \\
& PraNet & 0.712 & 0.640 & 0.699 & 0.820 & 0.847 & 0.872 & 0.043 \\
& MSEG & 0.735 & 0.666 & 0.724 & 0.834 & 0.859 & 0.875 & \underline{0.038}\\
& DCRNet & 0.704 & 0.631 & 0.684 & 0.821 & 0.840 & 0.848 & 0.052\\
& EU-Net & \underline{0.756} & \underline{0.681} & \underline{0.730} & 0.831 & 0.863 & 0.872 & 0.045\\
& SANet & 0.753 & 0.670 & 0.726 & \underline{0.837} & \underline{0.869} & \underline{0.878} & 0.043\\
& \textbf{Ours} & \textbf{0.801} & \textbf{0.717} & \textbf{0.770} & \textbf{0.869} & \textbf{0.902} & \textbf{0.916} & \textbf{0.030}\\
\hline
\multirow{10}{*}{ETIS} & UNet & 0.398 & 0.335 & 0.366 & 0.684 & 0.643 & 0.740 & 0.036 \\
& UNet++ & 0.401 & 0.344 & 0.390 & 0.683 & 0.629 & 0.776 & 0.035 \\
& SFA & 0.297 & 0.217 & 0.231 & 0.557 & 0.531 & 0.632 & 0.109 \\
& ACSNet & 0.578 & 0.509 & 0.530 & 0.754 & 0.737 & 0.764 & 0.059 \\
& PraNet & 0.628 & 0.567 & 0.600 & 0.794 & 0.808 & 0.841 & 0.031 \\
& MSEG & 0.700 & 0.630 & 0.671 & 0.828 & 0.854 & 0.890 & 0.015 \\
& DCRNet & 0.556 & 0.496 & 0.506 & 0.736 & 0.742 & 0.773 & 0.096 \\
& EU-Net & 0.687 & 0.609 & 0.636 & 0.793 & 0.807 & 0.841 & 0.067 \\
& SANet & \underline{0.750} & \underline{0.654} & \underline{0.685} & \underline{0.849} & \textbf{0.881} & \underline{0.897} & \textbf{0.015} \\
& \textbf{Ours} & \textbf{0.758} & \textbf{0.670} & \textbf{0.698} & \textbf{0.857} & \underline{0.868} & \textbf{0.906} & \underline{0.023} \\
\hline
\end{tabularx}
\end{table}

Our FLDNet consistently shows an impressive performance across different datasets, in several cases outperforming the other models by a significant margin. Particularly on the CVC-ColonDB dataset \cite{27}, FLDNet surpasses all other models, exhibiting a remarkable improvement in mean DICE (over 6

Apart from our own results, it is also notable that different models perform differently across various datasets. For example, the ACSNet model, which generally shows strong performance, struggles on the CVC-ColonDB dataset compared to other datasets. This could be due to the model's difficulty in handling the specific complexities presented in the CVC-ColonDB dataset. On the other hand, the MSEG model performs remarkably well on the ETIS dataset, where it provides highly competitive results, especially in terms of the $mE_{\xi}$ and $maxE_{\xi}$ metrics.

These observations demonstrate the uniqueness and challenges of each dataset and underline the importance of model versatility when tackling different types of data. It is precisely in this aspect that our FLDNet excels, exhibiting robust performance regardless of dataset complexity, thereby confirming its effectiveness for polyp segmentation.

\subsection{Visual Comparison }\label{sec:third_sub3}
Figure \ref{fig4} illustrates the prediction maps for polyp segmentation by different models. We have compared the prediction results of our proposed model with those of other models. The results reveal that our model significantly excels in emphasizing the polyp region and suppressing background noise, compared to other approaches. Especially when faced with challenging scenarios, our model can handle them adeptly and generate more accurate segmentation masks. These outcomes comprehensively demonstrate the significant superiority of our proposed model in polyp segmentation over other models.

\begin{figure}[t!]
\begin{center}
\includegraphics[width=0.7\textwidth]{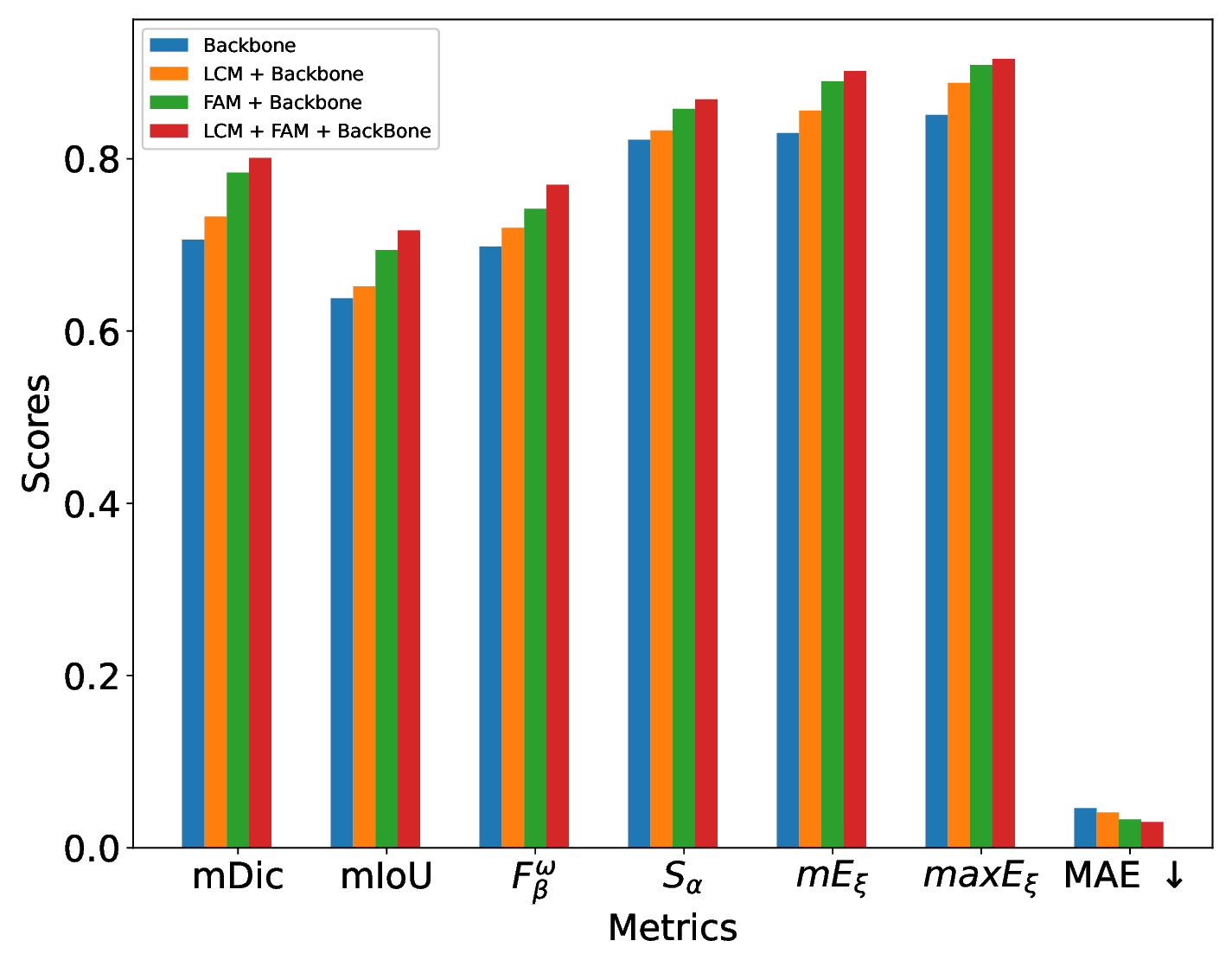}
\caption{Ablation Study Results of the different components of our model on different metrics.} \label{fig5}
\end{center}
\end{figure}

\subsection{Ablation Study}\label{sec:third_sub4}
To investigate the importance of each proposed component in the FLDNet, the ColonDB dataset was employed for ablation studies to validate the efficacy of the modules and to gain deeper insights into our model. 
As shown in Figure \ref{fig5}, all the modules or strategies are necessary for the final prediction. Combining all proposed methods, our model achieves state-of-the-art performance.

\section{Conclusion}
In this paper, we proposed a novel automatic polyp segmentation network, FLDNet. Extensive experiments have demonstrated that the proposed FLDNet achieved state-of-the-art performance on challenging datasets. The results of the ablation studies show that the proposed Local Context Module and Foreground-Aware Module significantly improved the segmentation performance of the model, alleviating the problem of difficulty in distinguishing polyps from surrounding mucosal environments. Furthermore, using PVT as a backbone endows our FLDNet with versatility, which can be further enhanced for polyp segmentation accuracy by adding different modules.In the future we will study the segmentation performance of the proposed FLDNet on more medical datasets, such as lung and MRI images to explore the generality of FLDNet.

%
%
%
%

\end{document}